\pdfoutput=1

\documentclass[11pt]{article}

\usepackage{emnlp2021}

\usepackage{times}
\usepackage{latexsym}

\usepackage[T1]{fontenc}

\usepackage[utf8]{inputenc}

\usepackage{microtype}

\usepackage{booktabs}
\usepackage{multirow}
\usepackage{amsmath}
\usepackage{amssymb}
\usepackage{xcolor}
\usepackage{pifont}

\usepackage{tikz}
\usepackage{subfigure}
\usepackage{pgfplots}
\usepackage{makecell,rotating,diagbox}
\usepackage{verbatim}
\usepackage{colortbl}

\usepackage{graphicx}

\usetikzlibrary{plotmarks}
\usetikzlibrary{patterns}
\usetikzlibrary{pgfplots.groupplots}

\usetikzlibrary{plotmarks}
\usetikzlibrary{arrows}
\usetikzlibrary{decorations}
\usetikzlibrary{decorations.pathreplacing}
\usetikzlibrary{decorations.markings}
\usetikzlibrary{backgrounds}
\usetikzlibrary{fit}
\usetikzlibrary{positioning}
\usetikzlibrary{calc}
\usetikzlibrary{patterns}
\usetikzlibrary{shadows}
\usetikzlibrary[shapes.multipart]
\usetikzlibrary{pgfplots.groupplots}

\definecolor{ugreen}{cmyk}{1,0,1,0.498}
\definecolor{lyyblue}{cmyk}{0.8278,0.3333,0,0.2941}
\definecolor{lyygreen}{cmyk}{0.6813,0,0.725,0.3725}
\definecolor{lyyred}{cmyk}{0,0.8855,0.8767,0.1098}
\definecolor{dblue}{cmyk}{1,0.5487,0,0.5569}
\definecolor{lypurple}{HTML}{e0c2c0}
\definecolor{lygreen}{HTML}{eff67b}
\definecolor{lyblue}{HTML}{d5ddef}
\definecolor{lyyellow}{HTML}{fdfab5}
\definecolor{lypink}{HTML}{ffe0db}
\definecolor{lyred}{HTML}{b71a3b}
\definecolor{lygrey}{HTML}{c4c0c2}
\definecolor{lyorange}{HTML}{eab586}

\newcommand{\tab}[1]{Table \ref{#1}}%
\newcommand{\fig}[1]{Fig. \ref{#1}}%
\title{Bag of Tricks for Optimizing Transformer Efficiency}

\author{
  Ye Lin\textsuperscript{1}\thanks{\ \ Authors contributed equally.}\ ,
  Yanyang Li\textsuperscript{2}$^{*}$,
  Tong Xiao\textsuperscript{1,3}\thanks{\ \ Corresponding author.},
  Jingbo Zhu\textsuperscript{1,3} \\
  \textsuperscript{1}NLP Lab, School of Computer Science and Engineering, \\
    Northeastern University, Shenyang, China \\
  \textsuperscript{2}The Chinese University of Hong Kong, Hong Kong, China \\
  \textsuperscript{3}NiuTrans Research, Shenyang, China \\
  {\tt \{linye2015,blamedrlee\}@outlook.com}\\
  {\tt \{xiaotong,zhujingbo\}@mail.neu.edu.cn} \\
}

\begin{document}
\maketitle
\begin{abstract}
Improving Transformer efficiency has become increasingly attractive recently. A wide range of methods has been proposed, e.g., pruning, quantization, new architectures and etc. But these methods are either sophisticated in implementation or dependent on hardware. In this paper, we show that the efficiency of Transformer can be improved by combining some simple and hardware-agnostic methods, including tuning hyper-parameters, better design choices and training strategies. On the WMT news translation tasks, we improve the inference efficiency of a strong Transformer system by 3.80$\times$ on CPU and 2.52$\times$ on GPU. The code is publicly available at https://github.com/Lollipop321/mini-decoder-network.
\end{abstract}

\section{Introduction}

\label{sec:introduction}

Standard implementation of Transformer \cite{DBLP:conf/nips/VaswaniSPUJGKP17} is not efficient for inference. Researchers have explored more efficient architectures \cite{DBLP:conf/acl/XiongZS18,DBLP:conf/ijcai/XiaoLZ0L19,DBLP:journals/corr/abs-2101-00542} or break the auto-regressive constraint in sequence generation \cite{DBLP:journals/corr/abs-1711-02281}. But most of these require significant updates of the model or hardware-dependent designs. It is still natural to ask whether the Transformer system can be optimized in a simple way \cite{DBLP:journals/corr/abs-2010-02416,DBLP:journals/corr/abs-2006-10369,DBLP:conf/emnlp/KimJHAHGB19,DBLP:conf/coling/WangT20a}.

In this paper we show that Transformer can be optimized for efficiency by a bag of techniques. These techniques are easy to implement and some of them have been tested in related studies. Here we focus on using them in combination for Transformer speedup which has not been well investigated. In particular, our work is based on the following facts:

\begin{itemize}
\item The default Byte-Pair Encoding (BPE) setting \cite{DBLP:conf/acl/SennrichHB16a} has a great impact on efficiency but is generally not optimal.
\item A shallow decoder (with a deeper encoder) is preferred for a fast system \cite{DBLP:journals/corr/abs-2006-10369}.
\item The attention model does not need to be multi-headed in some cases \cite{DBLP:conf/emnlp/BehnkeH20}.
\item The feedforward network sub-layer is removable \cite{DBLP:journals/corr/abs-2010-02416}.
\item Knowledge Distillation \cite{DBLP:journals/corr/HintonVD15} is crucial to squeeze out the last potential. Removing some regularization measures like label smoothing \cite{DBLP:conf/cvpr/SzegedyVISW16} also helps when training such models.

\end{itemize}

\begin{table}[t!]
  \centering
  \begin{tabular}{c|l|r|r}
  \hline
  \multicolumn{2}{c|}{\multirow{2}{*}{Module}} &
  \multicolumn{2}{c}{Time (s)} \\ 
  \cline{3-4}
  \multicolumn{2}{c|}{} &
  \multirow{1}{*}{Baseline} & 
  \multirow{1}{*}{MDN} \\
  \hline
  \multirow{2}{*}{Encoder} &
  \multirow{1}{*}{Attention} & 5.81 & 16.93 \\
  & \multirow{1}{*}{FFN} & 5.53 & 18.79 \\
  \hline
  \multirow{3}{*}{Decoder} &
  \multirow{1}{*}{Attention} & 223.55 & 8.74 \\
  & \multirow{1}{*}{FFN} & 38.26 & 0.00 \\
  & \multirow{1}{*}{Output} & 48.51 & 8.61 \\
  \hline
  \end{tabular}
  \caption{Profiling results of the Transformer baseline and our model on WMT14 En-De (FFN: the feedforward network, Output: the output projection).} 
  \label{tab:time}
\end{table}


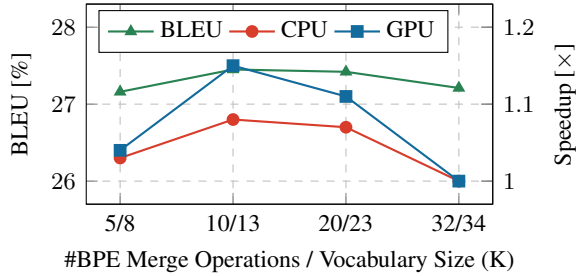
\begin{figure}[t!]
  \centering
  \begin{tikzpicture}
      \begin{axis}[
          width=0.9\linewidth,height=0.55\linewidth,
          yticklabel style={/pgf/number format/fixed,/pgf/number format/precision=1},
          ymin=26,ymax=28,
          enlarge y limits=0.15,
          ytick={26,27,28},
          ylabel={BLEU [\%]},
          ylabel near ticks,
          xlabel={\#BPE Merge Operations / Vocabulary Size (K)},
          xlabel near ticks,
          symbolic x coords={5/8,10/13,20/23,32/34},
          xmajorgrids=true,
          ymajorgrids=true,
          grid style=dashed,
          xtick=data,
          every tick label/.append style={font=\small},
          label style={font=\small},
        ]
          \addplot [lyygreen,thick,mark=triangle*] coordinates {
              (5/8,27.16) (10/13,27.45) (20/23,27.42) (32/34,27.21)
          };\label{bleuenc}
      \end{axis}
      \begin{axis}[
          width=0.9\linewidth,height=0.55\linewidth,
          yticklabel style={/pgf/number format/fixed,/pgf/number format/precision=1},
          ymin=1,ymax=1.2,
          enlarge y limits=0.15,
          xtick=\empty,
          ytick={1,1.1,1.2},
          ylabel={Speedup [$\times$]},
          ylabel near ticks,
          axis y line*=right,
          symbolic x coords={5/8,10/13,20/23,32/34},
          every tick label/.append style={font=\small},
          label style={font=\small},
          legend columns=3,
          legend pos=north west,
          legend style={font=\small},
        ]
          \addlegendimage{/pgfplots/refstyle=bleuenc}\addlegendentry{BLEU}
          \addplot [lyyred,thick,mark=*] coordinates {
              (5/8,1.03) (10/13,1.08) (20/23,1.07) (32/34,1.00) 
          };\addlegendentry{CPU}
          \addplot [lyyblue,thick,mark=square*] 
          coordinates {
              (5/8,1.04) (10/13,1.15) (20/23,1.11) (32/34,1.00) 
          };\addlegendentry{GPU}
      \end{axis}
  \end{tikzpicture}
  \caption{The number of BPE merge operations vs. BLEU and speedup on WMT14 En-De (Detailed setup could be found in Section \ref{sec:setup}).}
  \label{fig:bpe}
\end{figure}

All these methods are compatible with popular Transformer codebases. In this work, we implement them on the decoder side because it occupies the inference time in many sequence generation tasks \cite{DBLP:journals/corr/abs-2010-02416,DBLP:conf/emnlp/KimJHAHGB19}. The end result is a simplified and fast Transformer decoder (see \tab{tab:time}) - \emph{Mini-Decoder Network} (MDN). Experiments on the WMT14 En-De, WMT14 En-Fr and NIST12 Zh-En machine translation (MT) benchmarks demonstrate that the improved system achieves a 3.80$\times$ speedup on CPU and a 2.52$\times$ speedup on GPU with performance on par with the baseline. The speedup obtained is available on most modern hardware, as it does not depend on specific hardware or library, e.g., quantization \cite{DBLP:conf/emnlp/ChungKCKJP0L20} and unstructured pruning \cite{DBLP:journals/corr/abs-2102-00554} require the support of the latest hardware-dependent and acceleration libraries.

\section{Methods}

\label{sec:methods}

\subsection{Byte-Pair Encoding}

\label{sec:bpe}


Byte-Pair Encoding (BPE) \cite{DBLP:conf/acl/SennrichHB16a} breaks words into subword units. It starts from the alphabet and merges characters into the most frequent subword units, then segments words in sentences by these merged subword units. BPE reduces the risk of out-of-vocabulary words with a small vocabulary, but comes with the cost of longer sentences. If more merge operations are employed, the resulting sentences will be shorter, yet the vocabulary is larger as more subword units are presented. This leads to a tradeoff between sentence length and vocabulary size, and both of them have an impact on the efficiency. In \fig{fig:bpe}, the green line denotes the BLEU referencing the left axis. The red line denotes the speed change on CPU, the blue line denotes the speed change on GPU, they are both referencing the right axis. As shown in \fig{fig:bpe}, tuning this hyper-parameter provides a considerable speedup without loss in performance, though most previous work simply adopts the default setting (32K). In our experiments we choose 10K because it is sufficient for good performance.



\begin{figure}[t!]
  \centering
  \hspace*{\fill}
    \begin{tikzpicture}

        \tikzstyle{block} = [rectangle,minimum height=1.0cm,font=\small,rounded corners=3pt]
        \tikzstyle{encoder block} = [block,draw=black,fill=gray!10,minimum width=2cm]
        \tikzstyle{decoder block} = [block,draw=black,fill=gray!10,minimum width=4.5cm,minimum height=6.6cm]
        \tikzstyle{self attention} = [block,draw=black,fill=orange!10,minimum width=4cm,minimum height=2cm]
        \tikzstyle{cross attention} = [block,draw=black,fill=orange!10,minimum width=4cm,minimum height=2cm]
        \tikzstyle{attention} = [block,draw=black,fill=blue!15,minimum width=2cm,minimum height=0.6cm]
        \tikzstyle{attention1} = [block,draw=gray,fill=blue!15,fill opacity=0.1,minimum width=2cm,minimum height=0.6cm]
        \tikzstyle{ffn} = [block,draw=gray,fill=lyyblue!40,fill opacity=0.1,minimum width=4cm,minimum height=0.6cm]
        \tikzstyle{softmax} = [block,draw=gray,fill=lyygreen!40,fill opacity=0.1,minimum width=4cm,minimum height=0.6cm]
        \tikzstyle{softmax1} = [block,draw=gray,fill=lyygreen!40,minimum width=2cm,minimum height=0.6cm]

        \begin{pgfonlayer}{background}    
        \node[encoder block] (eb) at (0,0){};
        \node[decoder block,anchor=south west] (db) at ([xshift=0.5cm]eb.south east) {};
        \node[font=\footnotesize,align=center,anchor=center] () at (eb.center) {Transformer\\Encoder};
        \node[font=\footnotesize,align=center,anchor=north] (se) at ([yshift=-0.6cm]eb.south) {Source Embedding};
        \node[font=\footnotesize,align=center,anchor=north] (te) at ([yshift=-0.6cm]db.south) {Target Embedding};
        \node[font=\footnotesize,align=center,anchor=east] (ld) at ([yshift=0.2cm]db.west) {$1\times$};
        \end{pgfonlayer}
        
        \begin{pgfonlayer}{background}    
        \node[self attention,anchor=south] (sa) at ([yshift=0.4cm]db.south) {};
        \node[font=\footnotesize,align=center,anchor=north west] () at (sa.north west) {Self-Attention};
        \node[attention1,anchor=north east] (a1) at ([xshift=-0.3cm,yshift=-0.5cm]sa.north east) {};
        \node[attention1,anchor=north east] (a2) at ([xshift=-0.2cm,yshift=-0.1cm]a1.north east) {};
        \node[attention1,anchor=north east] (a3) at ([xshift=-0.2cm,yshift=-0.1cm]a2.north east) {};
        \node[attention1,anchor=north east] (a4) at ([xshift=-0.2cm,yshift=-0.1cm]a3.north east) {};
        \node[attention1,anchor=north east] (a5) at ([xshift=-0.2cm,yshift=-0.1cm]a4.north east) {};
        \node[attention1,anchor=north east] (a6) at ([xshift=-0.2cm,yshift=-0.1cm]a5.north east) {};
        \node[attention1,anchor=north east] (a7) at ([xshift=-0.2cm,yshift=-0.1cm]a6.north east) {};
        \node[attention,anchor=north east] (a8) at ([xshift=-0.2cm,yshift=-0.1cm]a7.north east) {};
        \node[font=\scriptsize,align=center,anchor=center] () at (a8.center) {Scaled Dot-Product\\Attention};
        \end{pgfonlayer}

        \begin{pgfonlayer}{background}    
          \node[cross attention,anchor=south] (ca) at ([yshift=0.8cm]sa.north) {};
          \node[font=\footnotesize,align=center,anchor=north west] () at (ca.north west) {Cross-Attention};
          \node[attention1,anchor=north east] (a1) at ([xshift=-0.3cm,yshift=-0.5cm]ca.north east) {};
          \node[attention1,anchor=north east] (a2) at ([xshift=-0.2cm,yshift=-0.1cm]a1.north east) {};
          \node[attention1,anchor=north east] (a3) at ([xshift=-0.2cm,yshift=-0.1cm]a2.north east) {};
          \node[attention1,anchor=north east] (a4) at ([xshift=-0.2cm,yshift=-0.1cm]a3.north east) {};
          \node[attention1,anchor=north east] (a5) at ([xshift=-0.2cm,yshift=-0.1cm]a4.north east) {};
          \node[attention1,anchor=north east] (a6) at ([xshift=-0.2cm,yshift=-0.1cm]a5.north east) {};
          \node[attention1,anchor=north east] (a7) at ([xshift=-0.2cm,yshift=-0.1cm]a6.north east) {};
          \node[attention,anchor=north east] (a8) at ([xshift=-0.2cm,yshift=-0.1cm]a7.north east) {};
          \node[font=\scriptsize,align=center,anchor=center] () at (a8.center) {Scaled Dot-Product\\Attention};
        \end{pgfonlayer}

        \begin{pgfonlayer}{background}    
          \node[ffn,anchor=south] (ffn) at ([yshift=0.4cm]ca.north) {};
          \node[font=\footnotesize,align=center,opacity=0.6,anchor=center] () at (ffn.center) {FFN};
        \end{pgfonlayer}

        \begin{pgfonlayer}{background}    
          \node[softmax,anchor=south] (softmax) at ([yshift=0.4cm]db.north) {};
          \node[softmax1,anchor=center] (softmax1) at (softmax.center) {};
          \node[font=\footnotesize,align=center,anchor=center] () at (softmax1.center) {Softmax};
          \node[font=\footnotesize,align=center,anchor=north] (op) at ([yshift=1.0cm]softmax1.north) {Output Probabilities};
        \end{pgfonlayer}

        \draw[-latex,thick] (se.north) to (eb.south);
        \draw[-latex,thick] (te.north) to (sa.south);
        \draw[-latex,thick] (ca.north) to (softmax.south);
        \draw[-latex,thick] (softmax.north) to (op.south);

        \draw[-latex,thick,rounded corners=3pt] (eb.north) |- ([yshift=0.4cm]sa.north) -- (ca.south);
        \draw[-latex,thick,rounded corners=3pt] (eb.north) |- ([xshift=-1cm,yshift=0.4cm]sa.north) -- ([xshift=-1cm]ca.south);
        \draw[-latex,thick,rounded corners=3pt] (sa.north) |- ([xshift=1cm,yshift=0.4cm]sa.north) -- ([xshift=1cm]ca.south);
        \draw[-latex,thick,rounded corners=3pt] (te.north) |- ([xshift=-1cm,yshift=-0.5cm]sa.south) -- ([xshift=-1cm]sa.south);
        \draw[-latex,thick,rounded corners=3pt] (te.north) |- ([xshift=1cm,yshift=-0.5cm]sa.south) -- ([xshift=1cm]sa.south);

    \end{tikzpicture}
  \hspace*{\fill}
  \caption{The model structure of our method (MDN).}
  \label{fig:mdn}
\end{figure}
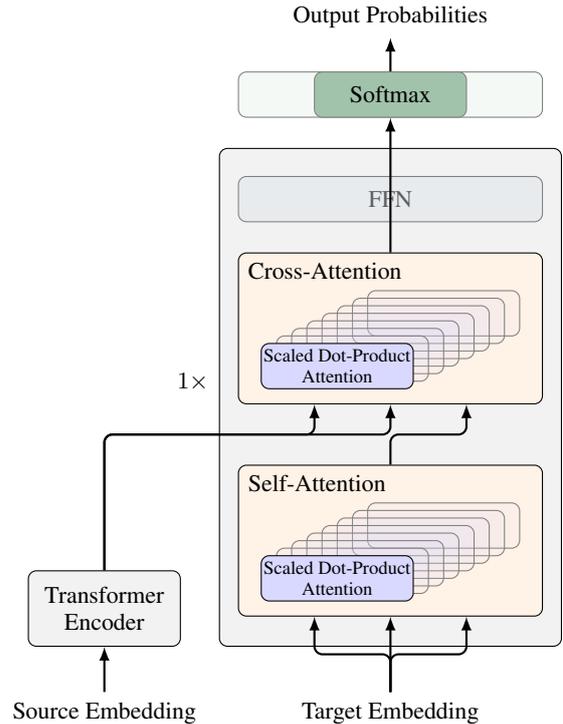

\subsection{Model Structure Updates}

\label{sec:model}

Inspired by the observations in \tab{tab:time}, the Transformer decoder can be improved for each of its components. 
In this section, we describe how to simplify Transformer in a systematic way as shown in \fig{fig:mdn}. \tab{tab:ablation_study_model} summarizes the contributions of each adopted method. We choose Baseline1 in \tab{tab:ablation_study_model} to analyze how each method in “Model Structure Updates” influences the model performance and inference speed after applying techniques from “Byte-Pair Encoding”.

\vspace{0.5em}
\noindent\textbf{Shallow Decoder.} Recent work has shown that the deep encoder and shallow decoder architecture is promising in system speedup \cite{DBLP:journals/corr/abs-2006-10369,DBLP:journals/corr/abs-2101-00542}. In this work we follow the same idea by restricting the decoder to a 1-layer network and stacking more encoder layers until the total number of parameters matches the baseline.

\vspace{0.5em}
\noindent\textbf{Pruning Heads.} Researchers have found that most heads could be safely pruned and leaving the performance intact \cite{DBLP:conf/acl/VoitaTMST19,DBLP:conf/nips/MichelLN19}. So we retain only one head in decoder attentions.

\vspace{0.5em}
\noindent\textbf{Dropping FFN.} \citet{DBLP:journals/corr/abs-2010-02416} suggests that FFN is the least important component in the decoder. So we drop all FFNs in the decoder. After dropping FFN, there are only attentions and no other non-linearity except layer normalization in the model.

\vspace{0.5em}
\noindent\textbf{Factorizing Output.} The weight matrix $W$ used in the output projection is significantly over-parameterized \cite{DBLP:conf/icml/GraveJCGJ17}, especially when other components are compressed. To address this, we employ the low-rank approximation \cite{DBLP:conf/iclr/LanCGGSS20} for this matrix $W=AB^T$ to help to reduce the computation cost, where $A \in \mathbb{R}^{V\times E}$, $B \in \mathbb{R}^{H\times E}$, $V$ is the vocabulary size, $E$ is the desired rank and $H$ is the hidden size. We choose $E=64$ in our experiments.

\begin{table}[!t]
  \centering
  \setlength{\tabcolsep}{1.3mm}{
    \begin{tabular}{l|c|r|r}
      \hline
      \multicolumn{1}{c|}{\multirow{2}{*}{System}}
      &
      \multicolumn{1}{c|}{\multirow{2}{*}{BLEU}} &
      \multicolumn{2}{c}{Speed (sent./s)} \\
      \cline{3-4}
      & &   
      \multicolumn{1}{c|}{CPU} & 
      \multicolumn{1}{c}{GPU} \\
      \hline
      \multirow{1}{*}{Baseline1} & 27.45 & 7.72 & 149.01 \\
      \hline
      \multirow{1}{*}{+ Shallow Decoder} & 26.46 & 19.44 & 247.04 \\
      \multirow{1}{*}{+ Pruning Heads} & 26.62 & 12.91 & 172.25 \\
      \multirow{1}{*}{+ Dropping FFN} & 26.91 & 8.58 & 189.86 \\
      \multirow{1}{*}{+ Factorizing Output} & 27.14 & 7.92 & 168.21 \\
      \hline
    \end{tabular}
  \caption{Results of adding each trick from Section \ref{sec:model} independently on WMT14 En-De (sent./s: translated sentences per second. Baseline1 is the baseline with 10K BPE merge operations).}
  \label{tab:ablation_study_model}
  }
\end{table}

\subsection{Training Strategies}

\label{sec:training}

In our work, the methods presented in Section \ref{sec:model} can make an extremely small decoder that contains only 0.3\% of the overall parameters. But we find that the model learned from scratch using the standard setting is much worse than the baseline (see \tab{tab:ablation_study_model}). For better training, some methods are necessary.
\tab{tab:ablation_study_training} illustrates how each proposed strategy contributes to reaching performance on par with the baseline. We choose Baseline2 in \tab{tab:ablation_study_training} to analyze how each method in “Training Strategies” influences the results after we have changed the BPE merge operations and simplified the model structure.

\vspace{0.5em}
\noindent\textbf{Deep Configuration.} Because our model is deep, we follow the deep model training setup provided in \citet{DBLP:conf/acl/WangLXZLWC19}.

\vspace{0.5em}
\noindent\textbf{Weight Distillation.} We also adopt a simplified version of weight distillation (WD) for training \cite{DBLP:journals/corr/abs-2009-09152}. This method initializes the student model with the corresponding weights from the teacher model, e.g., the first layer in the teacher encoder is reused in the first layer in the student encoder. Then it trains the student as in standard knowledge distillation \cite{DBLP:journals/corr/HintonVD15}. Since our encoder is much deeper than the baseline, we initialize it in a round-robin manner. For the decoder, we randomly select one head from the teacher model for initialization, and the low-rank approximation of output projection is initialized by the SVD result of the teacher output projection \cite{DBLP:books/ox/07/GolubR07}.

\vspace{0.5em}
\noindent\textbf{Weak Regularization.} Because our decoder is small, we do not need to impose a strong regularization on it. We remove the dropout in the decoder and label smoothing. Dropout and label smoothing indeed do not have impacts on the inference speed. But changing them will train different models, which are unlikely to have exactly the same behavior. So some deviations in the inference speed are expected.

\begin{table}[!t]
  \centering
  \setlength{\tabcolsep}{1.3mm}{
    \begin{tabular}{l|c|r|r}
      \hline
      \multicolumn{1}{c|}{\multirow{2}{*}{System}}
      &
      \multicolumn{1}{c|}{\multirow{2}{*}{BLEU}} &
      \multicolumn{2}{c}{Speed (sent./s)} \\
      \cline{3-4}
      & &   
      \multicolumn{1}{c|}{CPU} & 
      \multicolumn{1}{c}{GPU} \\
      \hline
      \multirow{1}{*}{Baseline2} & 22.83 & 23.19 & 291.00 \\
      \hline
      \multirow{1}{*}{+ Deep Configuration} & 23.78 & 24.21 & 305.03 \\
      \multirow{1}{*}{+ WD} & 26.97 & 24.56 & 352.09 \\ 
      \multirow{1}{*}{- Decoder Dropout} & 23.25 & 24.51 & 277.38 \\
      \multirow{1}{*}{- Label Smoothing} & 23.22 & 24.42 & 286.54 \\  
      \hline
    \end{tabular}
  \caption{Results of adding each trick from Section \ref{sec:training} independently on WMT14 En-De (sent./s: translated sentences per second. Baseline2 is the baseline with 10K BPE merge operations and tricks from Section \ref{sec:model}).}
  \label{tab:ablation_study_training}
  }
\end{table}

\begin{table*}[t!]
  \centering
  \begin{tabular}{c|l|c|c|r|c|r|c|r}
  \hline
  &
  \multicolumn{1}{l|}{System} &
  \multicolumn{1}{c|}{Test} &
  \multicolumn{1}{c|}{Valid} &
  \multicolumn{1}{c|}{Speed (CPU)} &
  Speedup &
  \multicolumn{1}{c|}{Speed (GPU)} &
  Speedup &
  \multicolumn{1}{c}{\#Params} \\
  \hline
  \multirow{6}{*}{\rotatebox{90}{En-De}} &
  \multirow{1}{*}{Baseline} & 27.21 & 25.53 & 7.17 sent./s & 1.00$\times$ & 129.02 sent./s & 1.00$\times$ & 96M \\
  & \multirow{1}{*}{+ BPE 10K} & 27.45 & 25.60 & 7.72 sent./s & 1.08$\times$ & 149.01 sent./s & 1.15$\times$ & 76M \\
  &  \multirow{1}{*}{AAN} & 27.05 & 25.11 & 9.82 sent./s & 1.37$\times$ & 147.11 sent./s & 1.14$\times$ & 96M \\
  &  \multirow{1}{*}{SAN} & 26.88 & 24.99 & 9.10 sent./s & 1.27$\times$ & 179.49 sent./s & 1.39$\times$ & 80M \\
  &  \multirow{1}{*}{CAN} & 27.20 & 25.23 & 8.00 sent./s & 1.12$\times$ & 279.39 sent./s & 2.17$\times$ & 100M \\
  \cline{2-9}
  &  \multirow{1}{*}{MDN} & 27.23 & 25.37 & 23.52 sent./s & 3.28$\times$ & 326.78 sent./s & 2.53$\times$ & 96M \\
  \cline{1-9}
  \multirow{6}{*}{\rotatebox{90}{En-Fr}} &
  \multirow{1}{*}{Baseline} & 40.70 & 46.74 & 5.80 sent./s & 1.00$\times$ & 130.56 sent./s & 1.00$\times$ & 111M \\
  & \multirow{1}{*}{+ BPE 10K} & 40.52 & 46.51 & 6.50 sent./s & 1.12$\times$ & 143.21 sent./s & 1.10$\times$ & 81M \\
  &  \multirow{1}{*}{AAN} & 40.44 & 46.43 & 8.00 sent./s & 1.38$\times$ & 140.66 sent./s & 1.08$\times$ & 96M \\
  &  \multirow{1}{*}{SAN} & 40.70 & 46.44 & 7.48 sent./s & 1.29$\times$ & 157.50 sent./s & 1.21$\times$ & 80M \\
  &  \multirow{1}{*}{CAN} & 40.16 & 46.40 & 6.40 sent./s & 1.10$\times$ & 194.67 sent./s & 1.49$\times$ & 100M \\
  \cline{2-9}
  &  \multirow{1}{*}{MDN} & 40.58 & 46.43 & 22.86 sent./s & 3.94$\times$ & 352.79 sent./s & 2.70$\times$ & 100M \\
  \cline{1-9}
  \multirow{6}{*}{\rotatebox{90}{Zh-En}} &
  \multirow{1}{*}{Baseline} & 45.84 & 50.98 & 4.09 sent./s & 1.00$\times$ & 90.87 sent./s & 1.00$\times$ & 102M \\
  & \multirow{1}{*}{+ BPE 10K} & 45.34 & 51.41 & 4.35 sent./s &1.06$\times$ & 91.79 sent./s & 1.01$\times$ & 79M \\
  &  \multirow{1}{*}{AAN} & 44.87 & 51.26 & 5.00 sent./s & 1.22$\times$ & 91.30 sent./s & 1.00$\times$ & 102M \\
  &  \multirow{1}{*}{SAN} & 44.82 & 50.73 & 4.99 sent./s & 1.22$\times$ & 123.18 sent./s & 1.36$\times$ & 102M \\
  &  \multirow{1}{*}{CAN} & 40.11 & 46.25 & 6.09 sent./s & 1.49$\times$ & 168.89 sent./s & 1.86$\times$ & 107M \\
  \cline{2-9}
  &  \multirow{1}{*}{MDN} & 44.51 & 51.43 & 17.07 sent./s & 4.17$\times$ & 211.31 sent./s & 2.33$\times$ & 99M \\
  \cline{1-9}
  \hline
  \end{tabular}
  \caption{Results on the WMT14 En-De and En-Fr tasks (sent./s: translated sentences per second).}
  \label{tab:main_result}
\end{table*}

\section{Experiments}

\subsection{Setup}

\label{sec:setup}

We evaluate our methods on the WMT14 En-De, WMT14 En-Fr and NIST12 Zh-En machine translation tasks.
We tokenize every sentence using a script from Moses and segment every word into subword units using BPE \cite{DBLP:conf/acl/SennrichHB16a}. The number of the BPE merge operations is set to 32K in the baseline and 10K for the target language in our model. In addition, we remove sentences with more than 250 subword units \cite{DBLP:conf/acl/XiaoZZL12}.

We choose Transformer-base \cite{DBLP:conf/nips/VaswaniSPUJGKP17} as our baseline. 
The hyper-parameters of the Mini-Decoder Network (MDN) are the same as the baseline except for those mentioned in Section \ref{sec:training}.
To produce consistent results for distillation, we choose the baseline with 10K BPE merges as the teacher model, which has the same vocabulary as MDN.
We also compare our system with some recent proposed fast Transformer variants, e.g., AAN \cite{DBLP:conf/acl/XiongZS18}, SAN \cite{DBLP:conf/ijcai/XiaoLZ0L19} and CAN \cite{DBLP:journals/corr/abs-2101-00542}. 
Their settings are followed from their papers.

We report case-sensitive tokenized BLEU scores.
For all experiments, we test on the model ensemble by averaging the last 5 checkpoints. 
For inference, we use a batch size of 64 and a beam width of 4.
All models are evaluated on the NVIDIA TITAN V GPU and Intel(R) Xeon(R) Gold 5118 CPU.

\subsection{Results}

\tab{tab:main_result} shows the results of various systems. 
In both tasks, our method (MDN) has nearly the same performance as the baseline, but its speed is 3.80$\times$ and 2.52$\times$ faster on average for CPU and GPU. 
We find the baseline with 10K BPE merges is about 1.09$\times$ faster than the original baseline but with a similar performance, which suggests this BPE hyper-parameter is far from optimal for the baseline.

As for the recent work, i.e., AAN, SAN and CAN, all of them achieve performance similar to the baseline and are faster than the baseline (1.24$\times$$\sim$1.32$\times$ speedup on CPU) as reported in their papers. But our method outperforms these methods and runs consistently faster (3.80$\times$ speedup on CPU).
Although the acceleration of our method in GPU (2.52$\times$ speedup) is not as obvious as it in CPU (3.80$\times$ speedup), it still outperforms CAN by 1.40$\times$, which is highly optimized for GPU.

\begin{table}[!t]
  \centering
  \setlength{\tabcolsep}{1.3mm}{
    \begin{tabular}{l|c|r|r}
      \hline
      \multicolumn{1}{c|}{\multirow{2}{*}{System}}
      &
      \multicolumn{1}{c|}{\multirow{2}{*}{BLEU}} &
      \multicolumn{2}{c}{Speed (sent./s)} \\
      \cline{3-4}
      & &   
      \multicolumn{1}{c|}{CPU} & 
      \multicolumn{1}{c}{GPU} \\
      \hline
      \multirow{1}{*}{Baseline} & 27.21 & 7.17 & 129.02 \\
      \hline
      \rowcolor{lyyellow}
      \multirow{1}{*}{+ Merge Operations} & 27.45 & 7.72 & 149.01 \\
      \rowcolor{lyblue}
      \multirow{1}{*}{+ Shallow Decoder} & 26.46 & 19.44 & 247.04 \\
      \rowcolor{lyblue}
      \multirow{1}{*}{+ Pruning Heads} & 24.47 & 21.99 & 243.42 \\
      \rowcolor{lyblue}
      \multirow{1}{*}{+ Dropping FFN} & 23.26 & 22.36 & 268.01 \\
      \rowcolor{lyblue}
      \multirow{1}{*}{+ Output Factorization} & 22.83 & 23.19 & 291.00 \\ 
      \rowcolor{lypink}
      \multirow{1}{*}{+ Deep Configuration} & 23.78 & 24.21 & 305.03\\
      \rowcolor{lypink}
      \multirow{1}{*}{+ WD} & 27.09 & 22.96 & 344.21\\
      \rowcolor{lypink}
      \multirow{1}{*}{- Decoder Dropout} & 27.18 & 23.53 & 334.40 \\
      \rowcolor{lypink}
      \multirow{1}{*}{- Label Smoothing} & 27.23 & 23.52 & 326.78 \\
      \hline
    \end{tabular}
  \caption{Ablation study on WMT14 En-De. The colors refer to \colorbox{lyyellow}{Byte-Pair Encoding}, \colorbox{lyblue}{Model Structure Updates} and \colorbox{lypink}{Training Strategies} (sent./s: translated sentences per second).}
  \label{tab:ablation_study}
  }
\end{table}

\section{Analysis}

\begin{table*}[t!]
  \centering
  \begin{tabular}{c|l|c|c|c|c|c|c}
  \hline
  &
  \multicolumn{1}{l|}{System} &
  \multicolumn{1}{c|}{Test} &
  \multicolumn{1}{c|}{Valid} &
  \multicolumn{1}{c|}{Speed(CPU)} &
  Speedup &
  \multicolumn{1}{c|}{Speed(GPU)} &
  Speedup \\
  \hline
  \multirow{3}{*}{\rotatebox{90}{deep}} &
  \multirow{1}{*}{Baseline} & 29.43 & 27.82 & 6.02 sent./s & 1.00$\times$ & 121.57 sent./s & 1.00$\times$ \\
  & \multirow{1}{*}{+ BPE 10K} & 29.67 & 27.68 & 6.19 sent./s & 1.03$\times$ & 126.06 sent./s & 1.04$\times$ \\
  &  \multirow{1}{*}{MDN} & 29.02 & 27.64 & 15.00 sent./s & 2.49$\times$ & 254.25 sent./s & 2.09$\times$ \\
  \cline{1-8}
  \hline
  \end{tabular}
  \caption{Results of Transformer-deep on WMT14 En-De (sent./s: translated sentences per second).}
  \label{tab:larger_networks_result}
\end{table*}

\subsection{Ablation Study}

\label{sec:ablation}

\tab{tab:ablation_study} summarizes and compares the contributions of each proposed tricks described in Section \ref{sec:methods}.
Each row of \tab{tab:ablation_study} is the result of applying the current trick to the system obtained in the previous row.
This way helps to illustrate the compound effect of these tricks.

We observe that using any structure simplification trick brings a significant performance degradation but a considerable speedup.
However, our training strategies can make up for the performance loss.
Among them, WD is most effective and has a boost of more than 3 BLEU points.
Interestingly, we find that pruning attention heads does not have any speedup on GPU.
This is because the parallelism of GPU decouples the computation cost of attention heads from the head number.

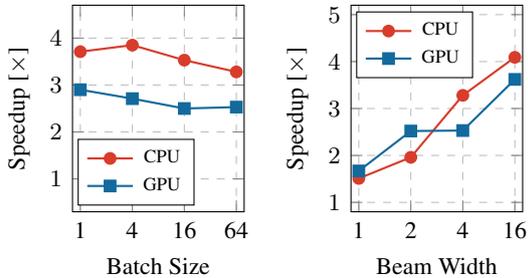
\begin{figure}[t!]
  \centering
  \begin{tikzpicture}
      \begin{groupplot}[
          group style={group size=3 by 1, horizontal sep=40pt},
          width=1.0\textwidth,
          height=0.3\textwidth,
          legend columns=1,
          legend pos=north west,
          legend style={font=\small},
          legend style={column sep=1.5pt},
      ]
        \nextgroupplot[
            width=0.24\textwidth,height=0.27\textwidth,
            yticklabel style={/pgf/number format/fixed,/pgf/number format/precision=1},
            ymin=0.5,
            ymax=4.5,
            ylabel={Speedup [$\times$]},
            ylabel near ticks,
            xlabel={Batch Size},
            xlabel near ticks,
            enlargelimits=0.05,
            symbolic x coords={1,4,16,64},
            xmajorgrids=true,
            ymajorgrids=true,
            legend pos=south west,
            grid style=dashed,
            xtick=data,
            every tick label/.append style={font=\small},
            label style={font=\small},
            ylabel style={yshift=0pt},            
            legend style={font=\scriptsize},
        ]
            \addplot [lyyred,thick,mark=*] 
            coordinates {
              (1,3.71) (4,3.85) (16,3.53) (64,3.28)
            };\addlegendentry{CPU}
            \addplot [lyyblue,thick,mark=square*] coordinates {
              (1,2.90) (4,2.71) (16,2.50) (64,2.53)
            };\addlegendentry{GPU}
        \nextgroupplot[
            width=0.24\textwidth,height=0.27\textwidth,
            yticklabel style={/pgf/number format/fixed,/pgf/number format/precision=1},
            ymin=1,
            ymax=5,
            ylabel={Speedup [$\times$]},
            ylabel near ticks,
            xlabel={Beam Width},
            xlabel near ticks,
            enlargelimits=0.05,
            symbolic x coords={1,2,4,16},
            xmajorgrids=true,
            ymajorgrids=true,
            grid style=dashed,
            xtick=data,
            every tick label/.append style={font=\small},
            label style={font=\small},
            ylabel style={yshift=0pt},            
            legend style={font=\scriptsize},
        ]
            \addplot [lyyred,thick,mark=*] 
            coordinates {
                (1,1.51) (2,1.96) (4,3.28) (16,4.09)
            };\addlegendentry{CPU}
            \addplot [lyyblue,thick,mark=square*] coordinates {
              (1,1.67) (2,2.52) (4,2.53) (16,3.62)
            };\addlegendentry{GPU}
      \end{groupplot}
  \end{tikzpicture}
  \caption{Batch size and beam width vs. speedup of MDN on WMT14 En-De.}
  \label{fig:sensitivity}
\end{figure}

\subsection{Experiments on Larger Networks}

\tab{tab:larger_networks_result} shows the results of the Transformer-deep model with a 48-layer encoder. The phenomenon here is similar to that in \tab{tab:main_result}. 
The acceleration on Transformer-deep is less obvious than on Transformer-base, as a deeper encoder consumes more inference time.
Moreover, compared with such a strong Transformer-deep teacher, MDN can still obtain a 2.49$\times$ speedup on CPU and a 2.09$\times$ speedup on GPU.

\subsection{Sensitivity Analysis}

We study the impact of two commonly tuned hyper-parameters at inference on our method, i.e., the batch size and the beam size. 
The left part of \fig{fig:sensitivity} shows that the speedup over the baseline decreases as the batch size increases, especially for GPU.
This is because our shallow decoder trick exploits the parallelism of the encoder for speedup, which is not available if the batch size is large.
This phenomenon is also observed by \citet{DBLP:journals/corr/abs-2101-00542}.
The right part of \fig{fig:sensitivity} shows that the speedup is more obvious with a larger beam width.
The reason is that the encoder occupies a larger portion of the inference cost at a small beam width and our work only save the cost of the decoder.

\begin{table}[!t]
  \centering
  \setlength{\tabcolsep}{3mm}{
    \begin{tabular}{l|c|r|r}
      \hline
      \multicolumn{1}{c|}{\multirow{2}{*}{System}} &
      \multicolumn{1}{c|}{\multirow{2}{*}{BLEU}} &
      \multicolumn{2}{c}{Speed (sent./s)} \\
      \cline{3-4}
      & &   
      \multicolumn{1}{c|}{CPU} & 
      \multicolumn{1}{c}{GPU} \\
      \hline
      \multirow{1}{*}{AAN} & 27.05 & 9.82 & 147.11 \\
      \hline
      \multirow{1}{*}{+ Ours} & 27.21 & 19.32 & 259.10 \\
      \hline
      \multirow{1}{*}{CAN} & 27.20 & 8.00 & 279.39 \\
      \hline
      \multirow{1}{*}{+ Ours} & 27.25 & 10.62 & 293.43 \\
      \hline
    \end{tabular}
  \caption{Combining the proposed tricks with AAN and CAN on WMT14 En-De (sent./s: translated sentences per second).}
  \label{tab:aan_can}
  }
\end{table}

\subsection{Combining with Other Models}

Our method is a bag of generic tricks and can be applied to other models.
We choose AAN and CAN for testing, because AAN runs the fastest on CPU and CAN runs the fastest on GPU according to \tab{tab:main_result}.
\tab{tab:aan_can} shows that both AAN and CAN benefit from techniques presented in this paper. Without loss in performance, AAN obtains a 1.97$\times$ speedup on CPU and CAN obtains a 1.05$\times$ speedup on GPU.
It shows that AAN and CAN eventually have a similar BLEU score and speed as MDN, indicating that a highly optimized Transformer baseline already a strong candidate by itself.

\section{Conclusion}

In this work, we present a bag of tricks to optimize the efficiency of the standard Transformer. The resulting model achieves a 3.61$\times$ speedup on CPU and a 2.62$\times$ speedup on GPU without loss in performance.

\section*{Acknowledgments}

This work was supported in part by the National Science Foundation of China (Nos. 61876035 and 61732005), the National Key R\&D Program of China (No. 2019QY1801), and the Ministry of Science and Technology of the PRC (Nos. 2019YFF0303002 and 2020AAA0107900). The authors would like to thank anonymous reviewers for their comments.

\bibliography{anthology,custom}
\bibliographystyle{acl_natbib}

\end{document}